\documentclass[11pt,a4paper]{article}
\usepackage[hyperref]{acl2017}
\usepackage{times}
\usepackage{latexsym}
\usepackage{amsmath,amsfonts,amssymb}
\usepackage{mathtools}
\usepackage{enumitem}
\usepackage{hyperref}
\usepackage[parfill]{parskip}
\usepackage{booktabs}
\usepackage{graphicx}
\usepackage{natbib}
\usepackage{xcolor}
\usepackage{url}
\usepackage{multicol}

\aclfinalcopy % Uncomment this line for the final submission
\defcitealias{dozat2017biaffine}{D \& M (2017)}

\title{Sequential Graph Dependency Parser}

\author{Sean Welleck\\
  New York University \\
  {\tt wellecks@nyu.edu} \\\And
  Kyunghyun Cho \\
  New York University \\
  CIFAR Azrieli Global Scholar \\
  Facebook AI Research \\
  {\tt kyunghyun.cho@nyu.edu} \\}

\date{}

\begin{document}
\maketitle
\begin{abstract}
We propose a method for non-projective dependency parsing by incrementally predicting a set of edges. 
Since the edges do not have a pre-specified order, we propose a set-based learning method. Our method blends graph, transition, and easy-first parsing, including a prior state of the parser as a special case. The proposed transition-based method successfully parses near the state of the art on both projective and non-projective languages, without assuming a certain parsing order.
\end{abstract}

\section{Introduction}

Dependency parsing methods can be categorized as graph-based 
and transition-based. Typical graph-based methods support non-projective parsing, but introduce independence assumptions and rely on external decoding algorithms. Conversely, transition-based methods model joint dependencies, but without modification are typically limited to projective parses.

There are two recent exceptions of interest here. \cite{ma2018stack} developed the Stack-Pointer parser, a transition-based, non-projective parser that maintains a stack populated in a top-down, depth-first manner, and uses a pointer network to determine the dependent of the stack's top node, resulting in a transition sequence of length $2n-1$. Recently, \cite{gonzalez2019left} developed a variant of the Stack-Pointer parser which parses in $n$ steps by traversing the sentence left-to-right, selecting the \textit{head} of the current node in the traversal, while incrementally checking for, and prohibiting, cycles. 

We take inspiration from both graph-based and transition-based approaches by viewing parsing as sequential graph generation. In this view, a graph is incrementally built by adding edges to an edge set. No distinction between projective and non-projective trees is necessary. Since edges do not have a pre-specified order, we propose a set-based learning method. Like \cite{gonzalez2019left}, our parser runs in $n$ steps. However, our learning method and transitions do not impose a left-to-right parsing order, allowing easy-first  \cite{tsuruoka2005bidirectional,goldberg2010efficient} behavior. Experimentally, we find that the proposed method can yield a sequential parser with preferred, input-dependent generation orders and performance gains over strong one-step methods.

\section{Graph Dependency Parser} 
Given a sentence $x=x_1,\ldots,x_N$, a dependency parser constructs a graph $G=(V,E)$ with $V=(x_0, x_1,\dots,x_N)$ and $E=\{(i,j)_1,\ldots (i,j)_{N}\}$, where $x_0$ is a special root node, and $E\subset\mathcal{E}$ forms a dependency tree.\footnote{See properties (1-5) in Appendix A.}

We describe a family of sequential graph-based dependency parsers. A parser in this family generates a \textit{sequence} of graphs where $V$ is fixed and $E=\bigcup_{t=1}^T E_t$:%
\begin{align}
    H^{\text{enc}}  &= f_{\text{enc}}(x_0,\ldots x_N)\\
    H^{\text{head}}_t, H^{\text{dep}}_t &= f_{\text{V}}(H^{\text{enc}}, E_{<t}, h_{t-1})\\
    S_t &= f_{\text{E}}(H^{\text{head}}_t, H^{\text{dep}}_t, S_{t-1})\\
    E_t &= f_{\text{dec}}(S_t, E_{<t}).
\end{align}
Steps (2-4) run for $T\leq N$ time-steps. At each time-step, first $f_V$ generates head and dependent representations for each vertex, $H_t^{\cdot}\in \mathbb{R}^{V\times d_H}$, based on vertex representations $H^{\text{enc}}\in \mathbb{R}^{V\times d}$, previously predicted edges $E_{<t}$, and a recurrent state $h_{t-1}\in \mathbb{R}^d$. Then $f_E$ computes a score for every possible edge, $S_t\in \mathbb{R}^{V\times V}$, and the scores are used by $f_{\text{dec}}$ to predict a set of edges $E_t$.

This general sequential family includes the biaffine parser of \cite{dozat2017biaffine} as a one-step special case, as well as a recurrent variant which we discuss below.

\subsection{Biaffine One-Step}\label{ssec:onestep} The Biaffine parser of \cite{dozat2017biaffine} is a one-step variant, implementing steps (1-4) using a bidirectional LSTM, head and dependent neural networks, a biaffine scorer, and a maximum spanning tree decoder, respectively:%
\begin{align*}
    H^{\text{enc}}&=\text{BiLSTM}(x_1,\ldots,x_N)\\
    H^{\text{head}},H^{\text{dep}} &= \text{MLP}^{\text{h}}(H^{\text{enc}}), \text{MLP}^{\text{d}}(H^{\text{enc}})\\
    S&=\text{BiAffine}(H^{\text{head}},H^{\text{dep}})\\
    E&=\text{MST}(S),
\end{align*}
where each row of scores $S^{(i)}$ is interpreted as a distribution over $i$'s potential head nodes:
\begin{align*}
    p((j\rightarrow i)|x) &\propto \text{softmax}_j(S^{(i)}),
\end{align*}
and $\text{MST}(\cdot)$ is an off-the-shelf maximum-spanning-tree algorithm. This model assumes conditional independence of the edges.

% -- old model that did not work as well:
% \subsection{Recurrent Hidden} 
% We propose a variant which iteratively adjusts a distribution over edges at each step, based on the predictions so far. It can model label dependencies since it adjusts scores as a function of previous predictions $E_t$. It uses recurrent and attentional contexts to adjust representations, namely: 
% \begin{align*}
%     f_{\text{update}}(E_{t-1}, H_{<t}) &\triangleq H_{t-1} + \tilde{c}_t\\
%     \text{where }\tilde{c}_t &= f_{\text{attn}}(H_{t-1}, c_t) + c_t,\\
%     c_t, h_t  &= \text{LSTM}(f_{\text{emb}}(E_{t-1}), h_{t-1}).
%     % e_t &= f_{\text{emb}}(E_t)
% \end{align*}
% %
% The resulting $H_t$ is used to adjust the previous scores, yielding a distribution over edges:
% \begin{align}\label{eq:dist}
%     f_{\text{score}}(S_{t-1}, H_t)&\triangleq S_{t-1} + \text{BiAffine}(H_t,H_t)\nonumber\\
%     p((i\rightarrow j)|\textbf{x})&\propto \text{softmax}(\text{flatten}(S_t)).
% \end{align}
%
% We use a bidirectional LSTM as $f_{\text{enc}}$.
%
\subsection{Recurrent Weight} \label{ssec:recurrent}
We propose a variant which iteratively adjusts a distribution over edges at each step, based on the predictions so far. 
A recurrent function generates a weight matrix $W$ which is used to form vertex embeddings and in turn adjust edge scores. 

Specifically, we first obtain an initial score matrix $S_0$ using the biaffine one-step parser (\ref{ssec:onestep}), and initialize a recurrent hidden state $h_0$ using a linear transformation of $f_{\text{enc}}$'s final hidden state. Then $f_{V}$ is defined as:
\begin{align*}
W, h_{t} &= \text{LSTM}(f_{\text{emb}}(E_{t-1}), h_{t-1})\\
H^{\text{head}}_t&=\text{emb}_h(0,\ldots,N)W\\
H^{\text{dep}}_t&=\text{emb}_d(0,\ldots,N)W,
\end{align*}
and $f_{\text{E}}(H^{\text{head}}_t,H^{\text{dep}}_t,S_{t-1})$ is defined as:
\begin{align*}
S_t^{\Delta} &= \text{BiAffine}(H^{\text{head}}_t, H^{\text{dep}}_t)\\
S_{t}&=S_{t-1}+S_t^{\Delta},
\end{align*}
where $t$ ranges from 1 to $N$,
$W\in \mathbb{R}^{d_{\text{emb}}\times d_H}$, 
and each $\text{emb}_{(\cdot)}:\mathbb{N}\rightarrow \mathbb{R}^{d_{\text{emb}}}$ is a learned embedding layer, 
yielding  $\text{emb}_{(\cdot)}(0,\ldots,N)$ in $\mathbb{R}^{V\times d_{\text{emb}}}$. 
We use a bidirectional LSTM as $f_{\text{enc}}$.

The scores at each step yield a distribution over all $V\times V$ edges, which we denote by $\pi$:
\begin{align}\label{eq:dist}
    \pi((i\rightarrow j)|E_{<t},x)&\propto \text{softmax}(\text{flatten}(S_t)).
\end{align}
Unlike the one-step model, this recurrent model can predict edges based on past predictions.
\paragraph{Inference} We must ensure the incrementally decoded edges $E=\bigcup_{t=1}^T E_t$ form a valid dependency tree. To do so, we choose $f_{\text{dec}}$ to be a decoder which greedily selects valid edges, 
\begin{align*}
    E_t &= f_{\text{valid}}(S_t,E_{<t}),
\end{align*}
which we refer to as the \textbf{valid decoder}, detailed in Appendix \ref{apx:decoder}. We only predict one edge per step $(|E_t|=1)$, leaving the setting of multiple predictions per step as future work.
\paragraph{Embedding Edges} We embed a predicted edge $E_t=\{\hat{(i,j)}\}$ as:
\begin{align*}
f_{\text{emb}}(E_{t}) &= e_{\text{edge}}; e_{\text{head}}; e_{\text{dependent}}\\
    e_{\text{edge}} &= W_eH^{\text{enc}}_{(i)}-W_eH^{\text{enc}}_{(j)}\\
    e_{\text{head}} &= \text{emb}_h(i)\\
    e_{\text{dependent}} &= \text{emb}_d(j),
\end{align*}
where $H^{\text{enc}}_{(\cdot)}\in \mathbb{R}^{d}$ are row vectors, $W_{e}\in\mathbb{R}^{d_e\times d}$ is a learned weight matrix, $\text{emb}_{(\cdot)}$ are learned embedding layers, and $;$ is concatenation. 
% --- multi-edge embedding (old):
% \paragraph{Embedding Predictions} We embed predicted edges $E_t=\{(i,j)_1,\ldots (i,j)_{|E_{t}|}\}$ as:
% \begin{align*}
% f_{\text{emb}}(E_{t}) &=\Phi\left(e((i,j)_1),\ldots, e((i,j)_{|E_{t}|})\right)\\
%     e(i,j) &= e_{\text{edge}}; e_{\text{head}}; e_{\text{dependent}}\\
%     e_{\text{edge}} &= W_eH^{\text{enc}}_{(i)}-W_eH^{\text{enc}}_{(j)}\\
%     e_{\text{head}} &= \text{emb}_h(i)\\
%     e_{\text{dependent}} &= \text{emb}_d(j),
% \end{align*}
% where $\Phi$ is a permutation-invariant function (or the identity when $|E_t|=1$), $H^{\text{enc}}_{(\cdot)}\in \mathbb{R}^{d}$ are row vectors, $W_{e}\in\mathbb{R}^{d_e\times d}$ is a learned weight matrix, $\text{emb}_{(\cdot)}$ are learned embedding layers, and $;$ is concatenation. %In our experiments, we assume $|E_t|=1$, and leave the setting of multiple predictions per step as future work.

\paragraph{Future Work} The proposed method does not specifically require a BiLSTM encoder, LSTM, or the BiAffine function. For instance,  $f_{\text{V}}$ could use a Transformer \cite{vaswani2017attention} to output states that are linearly transformed into $H^{\text{head}}$ and $H^{\text{dep}}$. 
Additionally, partial graphs $(V, E_{<t})$ might be embedded using neural networks specifically designed for graphs \cite{gilmer2017}.
Finally, predicting edge sets %$E_1,\ldots E_T$ 
of size greater than 1 could potentially be achieved using a partially-autoregressive model, trained with a `masked edges' objective, similar to recent work in machine translation with conditional masked language models \cite{ghazvininejad2019constant}. Each call to $f_{V}$ would involve a separate forward pass which calls a Transformer $f_{\text{enc}}$. The partial tree is encoded via non-masked inputs to $f_{\text{enc}}$. $f_{\text{E}}$ corresponds to having $V$ outputs, each a distribution over $V$ edges. The multi-step decoder (Appendix \ref{apx:decoder}) might be used at test time.

\section{Learning}
In this paper, we restrict to the case of predicting a single edge $\hat{(i,j)}$ per step, so that the recurrent weight model generates a sequence of edges with the goal of matching a target edge set, i.e. $\bigcup_{t=1}^{N}\hat{(i,j)}_t=E$. Since the target edges $E$ are a set, the model's generation order is not determined \textit{a priori}. As a result, we propose to use a learning method that does not require a pre-specified generation order and allows the model to learn input-dependent orderings. 

Our proposed method is based on the multiset loss \cite{welleck2018loss} and its recent extensions for non-monotonic generation \cite{welleck2019nonmonotonic}. The method is motivated from the perspective of learning-to-search \cite{daume2009searn,chang2015learning}, which involves learning a \textit{policy} $\pi_{\theta}$ that mimics an \textit{oracle policy} $\pi^*$. The policy maps \textit{states} to distributions over \textit{actions}. 

For the proposed graph parser, an action is an edge $(i,j)\in \mathcal{E}$, and a state $s_t$ is an input sentence $x$ along with the edges predicted so far, $\hat{E}_{<t}$. The policy is a conditional distribution over $\mathcal{E}$, $$\pi_{\theta}((i,j)|\hat{E}_{<t}, x),$$ such as the distribution in equation \eqref{eq:dist}.

Learning consists of minimizing a cost, computed by first sampling states from a \textit{roll-in} policy $\pi^{\text{in}}$, then using a \textit{roll-out policy} $\pi^{\text{out}}$ to estimate cost-to-go for all actions at the sampled states. Formally, we minimize the following objective with respect to $\theta$:
\begin{align}\label{eq:objective}
    \mathbb{E}_{x\sim \mathcal{D}}\mathbb{E}_{s_1,\ldots,s_{|x|}\sim \pi^{\text{in}}} \mathcal{C}(\pi_{\theta},\pi^{\text{out}},s_t).
\end{align}
This objective involves sampling a sentence $x$ from a dataset, sampling a sequence of edges from the roll-in policy, then computing a cost $\mathcal{C}$ at each of the resulting states. We now describe our choices of $\mathcal{C}$, $\pi^{\text{out}}$, $\pi^*$, and $\pi^{\text{in}}$, and evaluate them later in the experiments (\ref{sec:exprs}).

\subsection{Cost Function and Roll-Out} Following \cite{welleck2018loss,welleck2019nonmonotonic} we use a KL-divergence cost:
\begin{align}
    \mathcal{C}(\pi_{\theta},\pi^{\text{out}},s)&=D_{\text{KL}}(\pi^{\text{out}}(\cdot|s)|| \pi_{\theta}(\cdot|s)).
\end{align}
We use the oracle $\pi^*$ as the roll-out $\pi^{\text{out}}$.

\subsection{Oracle} Based on the free labels set in \cite{welleck2018loss}, we first define a \textit{free edge set} containing the un-predicted target edges at time $t$:
\begin{align}
    E_{\text{free}}^t &= E\ \backslash \bigcup_{t'=1}^{t-1} \hat{(i,j)}_{t'},
\end{align}
where $E_{\text{free}}^0=E$.
We then construct a family of oracle policies that place non-zero probability mass only on free edges:
\begin{align}
    \pi^*((i,j)|E_{\text{free}}^t)= \begin{cases}
    p_{ij} & (i,j)\in E_{\text{free}}^t\\
    0 & \text{otherwise.}
    \end{cases}
\end{align}
We now describe several oracles by varying how $p_{ij}$ is defined.
\paragraph{Uniform} This oracle treats each permutation of the target edge set as equally likely by assigning a uniform probability to each free edge:
\begin{align*}
    \pi^*_{\text{unif}}((i,j)|E_{\text{free}}^t)=\begin{cases}
    \frac{1}{|E_{\text{free}}^t|} & (i,j)\in E_{\text{free}}^t\\
    0 & \text{otherwise.}
    \end{cases}
\end{align*}

\paragraph{Coaching} Motivated by \cite{he2012imitation,welleck2019nonmonotonic}, we define a coaching oracle which weights free edges by $\pi_{\theta}$:
\begin{align*}
    \pi^*_{\text{coaching}}((i,j)|E_{\text{free}}^t)\propto \pi^*_{\text{unif}}(\cdot|E_{\text{free}}^t)\pi_{\theta}(\cdot|E_{<t}, X).
\end{align*}
This oracle prefers certain edge permutations over others, reinforcing $\pi_{\theta}$'s preferences. The coaching and uniform oracles can be mixed to ensure each free edge receives probability mass:
\begin{align}\label{eqn:betacoach}
    \beta \pi^*_{\text{unif}} + (1-\beta)\pi^*_{\text{coaching}},
\end{align}
where $\beta\in [0,1].$

\paragraph{Annealed Coaching} This oracle begins with the uniform oracle, then anneals towards the coaching oracle as training progresses by annealing the $\beta$ term in \eqref{eqn:betacoach}. This may prevent the coaching oracle from reinforcing sub-optimal permutations early in training.

\paragraph{Linearized} This oracle uses a deterministic function to linearize an edge set $E$ into a sequence $E_{\text{seq}}$. The oracle selects the $t$'th element of $E_{\text{seq}}$ at time $t$ with probability 1.  
We linearize an edge set in increasing edge-index order: $(i_1,j_1)$ precedes $(i_2,j_2)$ if $(i_1,j_1)<(i_2,j_2)$. This oracle serves as a baseline that is analogous to the fixed generation orders used in conventional parsers.

\subsection{Roll-In} The roll-in policy determines the state distribution that $\pi_{\theta}$ is trained on, which can address the mismatch between training and testing state distributions \cite{ross2011reduction,chang2015learning} or narrow the set of training trajectories. We evaluate several alternatives:
\begin{enumerate}
    \item \textbf{uniform} $(i,j)\sim \pi^*_{\text{unif}}$
    \item \textbf{coaching} $(i,j)\sim \pi_{\theta}\odot \pi^*_{\text{unif}}$
    \item \textbf{valid-policy} $(i,j)\sim \text{valid}(\pi_{\theta})$
\end{enumerate}
where $\text{valid}(\pi_{\theta})$ is the set of edges that keeps the predicted tree as a valid dependency tree. The coaching and valid-policy roll-ins choose edge permutations that are preferred by the policy, with valid-policy resembling test-time behavior.

\section{Experiments}\label{sec:exprs}
In Experiments \ref{ssec:expr1} and \ref{ssec:expr2} we evaluate on English, German, Chinese, and Ancient Greek since they vary with respect to projectivity, size, and performance in \cite{qi2018universal}. Based on these development set results, we then test our strongest model on a large suite of languages (\ref{ssec:expr4}).

\paragraph{Experimental Setup} Experiments are done using datasets from the CoNLL 2018 Shared Task \cite{zeman2018conll}. We build our implementation from the open-source version of \cite{qi2018universal}\footnote{\url{https://github.com/stanfordnlp/stanfordnlp}.}, and use their experimental setup (e.g. pre-processing, data-loading, pre-trained vectors, evaluation) which follows the shared task setup. Our model uses the same encoder from \cite{qi2018universal}. For the \cite{qi2018universal} baseline, we use their pretrained models\footnote{\url{https://stanfordnlp.github.io/stanfordnlp/installation_download.html}.} and evaluation script. For the \cite{dozat2017biaffine} baseline, we use the \cite{qi2018universal} implementation with auxiliary outputs and losses disabled, and train with the default hyper-parameters and training script.  For our models only, we changed the learning rate schedule (and model-specific hyper-parameters), after observing diverging loss in preliminary experiments with the default learning rate. Our models did not require the additional AMSGrad technique used in \cite{qi2018universal}. We evaluate validation UAS every 2k steps (vs. 100 for the baseline). Models are trained for up to 100k steps, and the model with the highest validation unlabeled attachment score (UAS) is saved. 
\begin{figure}[t]
\includegraphics[width=\columnwidth]{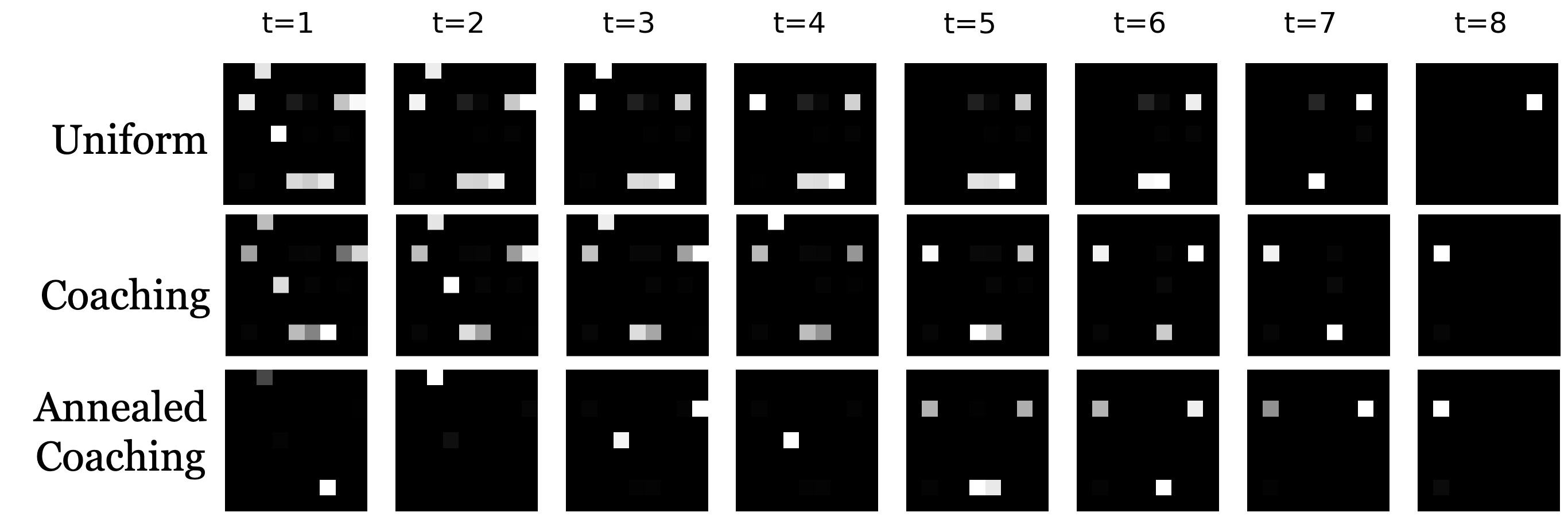}
\caption{Per-step edge distributions from recurrent weight models trained with the given oracle.}\label{fig:perstep}
\end{figure}

\subsection{Multi-Step Learning} \label{ssec:expr1} In this experiment we evaluate the sequential aspect of the proposed recurrent model by comparing it with one-step baselines. We compare against a baseline (`One-Step') that simply uses the first step's score matrix $S_0$ from the recurrent weight model and minimizes \eqref{eq:objective} for one time-step using a uniform oracle. At test time the valid decoder uses $S_0$ for all timesteps. We also compare against the biaffine one-step model of \cite{dozat2017biaffine} which uses Chu-Liu-Edmonds maximum spanning tree decoding instead of valid decoding. Since we only evaluate UAS, we disable its edge label output and loss. Finally, we compare against \cite{qi2018universal} which is based on \cite{dozat2017biaffine} plus auxiliary losses for length and linearization prediction.

% Experiment 1
\begin{table}[t]
%   \footnotesize
  \centering
  \setlength\tabcolsep{4.0pt}
\begin{tabular}{@{}l|cccc}
	\toprule
	 & \textbf{En} & \textbf{De} & \textbf{Grc} & \textbf{Zh} \\ 
	\midrule
	\citetalias{dozat2017biaffine} & 91.14 & 90.38 & 78.99& 86.50  \\
	\citet{qi2018universal}& 92.11 & 89.46 & 81.35 & 86.73  \\ % (their pretrained models)
	One-Step     & 91.74 & 91.07 & 79.60 & 86.61  \\
	Recurrent (U) & 91.92 & 91.02 & 79.15 & 86.69 \\
% 	Recurrent W. (C) & 92.03 & 91.16 & 79.96 & 86.37 \\  % stochastic coaching roll-in results 
	Recurrent (C) & 91.99 & 91.19 & 79.93 & 86.77 \\  % valid-greedy roll-in results
	\bottomrule
\end{tabular}%
\caption{\label{tbl:expr1} Development set UAS for single vs. multi-step methods. (U) is uniform oracle and roll-in, (C) is coaching with greedy valid roll-in ($\beta=0.5$). \citetalias{dozat2017biaffine} is an abbreviation for \cite{dozat2017biaffine}.} 
\vspace{-1em}
\end{table}

Results are shown in Table \ref{tbl:expr1}, including results for a recurrent model trained with coaching (`Recurrent (C)') using a mixture (eq. \ref{eqn:betacoach}) with $\beta=0.5$. The one-step baseline is strong, even outperforming the uniform recurrent variant on some languages. The recurrent weight model with coaching, however, outperforms the one-step and \cite{dozat2017biaffine} baselines on all four languages. Adding in auxiliary losses to the \cite{dozat2017biaffine} model yields improved UAS as seen in the \cite{qi2018universal} performance, suggesting that our proposed recurrent model might be improved further with auxiliary losses.

\paragraph{Temporal Distribution Adjustment} Figure \ref{fig:perstep} shows per-step edge distributions on an eight-edge example. The recurrent weight variants learned to adjust their distributions over time based on past predictions. The model trained with the uniform oracle has a decreasing number of high probability edges per step since it aims to place equal mass on each free edge $(i,j)\in \hat{E}^t_{\text{free}}$. The model trained with coaching learned to prefer certain free edges over others, but with $\beta=0.5$ the uniform term in the loss still encourages placing mass on multiple edges per step. By annealing $\beta$, however, the coaching model exhibits vastly different behavior than the uniform-trained policy. The low entropy distributions at early steps followed by higher entropy distributions later on (e.g. $t\in\{5,6\}$) may indicate easy-first behavior.% (taking entropy as an indicator of difficulty).

% Experiment 3
\begin{table}[t]
  \centering
\begin{tabular}{@{}ll|cc}
	\toprule
	\textbf{Oracle} & \textbf{Roll-in} & \textbf{UAS} & \textbf{Loss}\\ 
% 	\midrule
% 	H    & U & Oracle          & 90.75 & 1.28 \\   % Recurrent Hidden results (not included, per experiment 1)
% 	H    & C & Policy (Greedy) & 90.89 & 0.43 \\
% 	H    & C & Policy (Stoch.) & 90.89 & 0.41 \\
% 	H    & C & Policy-Oracle (Greedy) & 90.70 & 1.44 \\
% 	H    & C & Policy-Oracle (Stoch.) & 90.86 & 0.82 \\
% 	H    & C & Policy-Valid (Greedy) & 90.92 & 1.25 \\
% 	H    & C & Policy-Valid (Stoch.) & 90.81 & 1.12 \\
	\midrule
	Linear & $\pi^*_{\text{linear}}$   & 81.03 & 0.04\\
	\midrule
	U & $\pi^*_{\text{unif}}$   & 91.02 &  0.35\\
	C & $\pi^*_{\text{unif}}$   & 91.04 & 0.17     \\
% 	\midrule
% 	U & $\pi_{\theta}$   & 90.82 & 0.52\\
% 	C & $\pi_{\theta}$   & 91.02 & 0.30\\
% 	CA & $\pi_{\theta}$  & 90.94 & 0.25 \\
	\midrule
% 	U  & $p_{\theta}$ (Stoch.) & 91.14 & 0.50 \\
% 	C  & $p_{\theta}$ (Stoch.) & 91.05 & 0.27 \\
% 	CA & $p_{\theta}$ (Stoch.) & 91.12 &  0.20\\
% 	\midrule
	U & $\pi^{*}_{\text{coach}}$  & 90.93 & 0.45 \\
	C & $\pi^{*}_{\text{coach}}$  & 91.17 & 0.33 \\
	CA &$\pi^{*}_{\text{coach}}$  & 90.89 & 0.34 \\
	\midrule
% 	U & $p_{\theta*}$ (Stoch.) & 91.00 &  0.34\\
% 	C & $p_{\theta*}$ (Stoch.) & 91.16 & 0.23\\
% 	CA & $p_{\theta*}$ (Stoch.) & 90.94 & 0.06\\
% 	\midrule
	U & $\pi_{\theta}^{\text{valid}}$  & 90.99 &  0.51\\
	C & $\pi_{\theta}^{\text{valid}}$  &\textbf{91.19} & 0.31\\
	CA & $\pi_{\theta}^{\text{valid}}$  & 90.91 & 0.30 \\
% 	\midrule
% 	U & $p_{\theta\text{valid}}$ (Stoch.) & 90.97 & 0.50\\
% 	C & $p_{\theta\text{valid}}$ (Stoch.) & 91.09 & 0.28\\
% 	CA & $p_{\theta\text{valid}}$ (Stoch.) & 90.87 & 0.23\\
	\bottomrule
\end{tabular}%
\caption{\label{tbl:expr23} Varying oracle and roll-in policies on German. (U), (C), (A) refer to uniform, coaching, and annealing, respectively. The $\pi^{*}_{\text{coach}}$ and $\pi_{\theta}^{\text{valid}}$ roll-ins are mixtures with a uniform oracle, with $\beta=0.5$ for coaching (C), and $\beta$ linearly annealed by 0.02 every 2000 steps for annealing (CA).}
\vspace{-1em}
\end{table}

\subsection{Oracle and Roll-In Choice} \label{ssec:expr2} In this experiment, we study the effects of varying the oracle and roll-in distributions. 
Table \eqref{tbl:expr23} shows results on German, analyzed below. Models trained with coaching (C) use a mixture with $\beta=0.5$, after observing lower UAS in preliminary experiments with lower $\beta$. 
The $\pi^*_{\text{coach}}$ and $\pi^{\text{valid}}_{\theta}$ roll-ins use a mixture with $\beta=0.5$ and greedy decoding, which generally outperformed stochastic sampling.

\paragraph{Set-Based Learning}
The model trained with the linearized oracle (UAS 81.03), which teaches the model to adhere to a pre-specified generation order, significantly under-performs the set-based models (UAS $\geq 90.89$), which do not have a pre-specified generation order and can in principle learn strategies such as easy-first.

\paragraph{Coaching} Models trained with coaching (C, UAS $\geq 91.04$) had higher UAS and lower loss than models trained with the uniform oracle (U, UAS $\leq 91.02$), for all roll-in methods.  This suggests that for the proposed model, weighting free edges in the loss based on the model's distribution is more effective than a uniform weighting. 

Annealing the $\beta$ parameter generally did not further improve UAS (CA vs. C), possibly due to the annealing schedule or overfitting; despite lower losses with annealing, eventually validation UAS \textit{decreased} as training progressed. 

\paragraph{Roll-In} With the coaching oracle (C), the choice of roll-in impacted UAS, with coaching roll-in ($\pi^{*}_{\text{coach}}$, 91.17) and valid roll-in ($\pi_{\theta}^{\text{valid}}, 91.19)$ achieving higher UAS than uniform oracle roll-in $(\pi^*_{\text{unif}}, 91.04)$. This suggests that when using coaching, narrowing the set of training trajectories to those preferred by the policy may be more effective than sampling uniformly from the set of all correct trajectories. Based on these results, we use the coaching oracle and valid roll-in for training our final model in the next experiment.

\subsection{CoNLL 2018 Comparison} \label{ssec:expr4} In this experiment, we evaluate our best model on a diverse set of multi-lingual datasets. We use the CoNLL 2018 shared task datasets that have at least 200k examples, along with the four datasets used in the previous experiments. We train a recurrent weight model for each dataset using the coaching oracle and valid roll-in. We compare against \cite{qi2018universal} which placed highly in the CoNLL 2018 competition, reporting test UAS evaluated using their pre-trained models.

Table \ref{tbl:expr4} shows the results on the 19 datasets from 17 different languages. The proposed model trained with coaching achieves a higher UAS than the \citet{qi2018universal} model on 12 of the 19 datasets, plus two ties. % though many of the differences are small
%
% Experiment 4
\begin{table}[t]
%   \footnotesize
  \centering
\begin{tabular}{@{}l|ccc}
	\toprule
	 & \textbf{Ours} & \textbf{\citet{qi2018universal}} \\ 
	\midrule
	\textbf{AR}         & 88.22 & \textbf{88.35} \\
	\textbf{CA}         & \textbf{94.13} & \textbf{94.13} \\
	\textbf{CS (CAC)}   & \textbf{93.53} & 93.22 \\
	\textbf{CS (PDT)}   & \textbf{93.80} & 93.21 \\
	\textbf{DE}         & \textbf{88.39} & 87.21 \\
	\textbf{EN (EWT)}   & \textbf{91.28} & 91.21 \\
	\textbf{ES}         & \textbf{93.70} & 93.38 \\
	\textbf{ET}         & \textbf{89.56} & 89.40 \\
	\textbf{FR (GSD)}       & \textbf{91.07} & 90.90 \\
	\textbf{GRC (Perseus)}  & 80.90 & \textbf{82.77} \\
	\textbf{HI} & \textbf{96.78} & \textbf{96.78} \\
	\textbf{IT (ISDT)}      & 94.06 & \textbf{94.24} \\
	\textbf{KO (KAIST)}     & \textbf{91.02} & 90.55 \\
	\textbf{LA (ITTB)}      & \textbf{93.66} & 93.00 \\
	\textbf{NO (Bokmaal)}   & \textbf{94.63} & 94.27 \\
	\textbf{NO (Nynorsk)}   & \textbf{94.44} & 94.02 \\
	\textbf{PT}             & 91.22 & \textbf{91.67} \\
	\textbf{RU (SynTagRus)}   & \textbf{94.57} & 94.42 \\
	\textbf{ZH}   & 87.31 & \textbf{88.49} \\
	\bottomrule
\end{tabular}%
\caption{\label{tbl:expr4} Test set results (UAS) on datasets from the CoNLL 2018 shared task with greater than 200k examples, plus the Ancient Greek (GRC) and Chinese (ZH) datasets. Bold denotes the highest UAS on each dataset.}
\vspace{-1em}
\end{table}

\section{Related Work} 

Transition-based dependency parsing has a rich history, with methods generally varying by the choice of transition system and feature representation. Traditional stack-based arc-standard and arc-eager \cite{yamada03statistical, nivre2003efficient} transition systems only parse projectively, requiring additional operations for pseudo-non-projectivity \cite{gomez2014polynomial} or projectivity \cite{nivre2009nonprojective}, while list-based non-projective systems have been developed \cite{nivre2008algorithms}. Recent variations assume a generation order such as top-down \cite{ma2018stack} or left-to-right \cite{gonzalez2019left}. Other recent models focus on unsupervised settings \cite{kim2019urnng}. Our focus here is a non-projective transition system and learning method which does not assume a particular generation order.

A separate thread of research in sequential modeling has demonstrated that generation order can affect performance \cite{vinyals2015order}, both in tasks with set-structured outputs such as objects \cite{welleck2017saliency,welleck2018loss} or graphs \cite{li2018DGMG}, and in sequential tasks such as language modeling \cite{ford2018importance}. Developing models with relaxed or learned generation orders has picked up recent interest \cite{welleck2018loss,welleck2019nonmonotonic,gu2019insertion,stern2019insertion}. We investigate this for dependency parsing, framing the problem as sequential set generation without a pre-specified order.

Finally, our work is inspired by techniques for improving upon maximum likelihood training through error exploration and dynamic oracles \cite{goldberg2012dynamic,goldberg2013training}, and related techniques in imitation learning for structured prediction \cite{daume2009searn,ross2011reduction,he2012imitation,goodman2016noise}. In particular, our formulation is closely related to the framework of \cite{chang2015learning}, where our oracle can be seen as an optimal roll-out policy which computes action costs without explicit roll-outs.
\vspace{-1.2em}
\section{Conclusion} \vspace{-1.2em}We described a family of dependency parsers which construct a dependency tree by generating a sequence of edge sets, and a learning method that does not presuppose a generation order. Experimentally, we found that a `coaching' method, which weights actions in the loss according to the model, improves parsing accuracy compared to a uniform weighting and allows the parser to learn preferred, input-dependent generation orders. The model's sequential aspect, along with the coaching method and training on a state distribution which resembles the model's own behavior, yielded improvements in unlabeled dependency parsing over strong one-step baselines.

\section*{Acknowledgements}
This work was partly supported by STCSM 17JC1404100/1.
% 
% \newpage
\bibliography{acl2018}
\bibliographystyle{acl_natbib}

\appendix
\section{Sequential Valid Decoder}\label{apx:decoder} 
We wish to sequentially sample $E_1,E_2,\ldots,E_T$ from score matrices $S_1,S_2,\ldots,S_T$, respectively, such that $E=\bigcup_t E_t$ is a dependency tree. A dependency tree must satisfy:
\begin{enumerate}
    \item The root node has no incoming edges.
    \item Each non-root node has exactly one incoming edge.
    \item There are no duplicate edges.
    \item There are no self-loops.
    \item There are no cycles.
\end{enumerate}
We first consider predicting one edge per step $|E_t|=1$, then address the case $|E_t|\geq 1$.

\paragraph{One Edge Per Step} Let $x=x_0,x_1,\ldots,x_N$ where $x_0$ is a root node. We define a function $f_{\text{valid}}(S_t, E_{<t})\rightarrow (i,j)$ which chooses the highest scoring edge $(i,j)$ such that $E_{<t}\cup \{(i,j)\}$ is a dependency tree, given edges $E_{<t}$ and scores $S_t$. We represent $E_{<t}$ as an adjacency matrix $A_{<t}$, and implement $f_{\text{valid}}(S_t,A_{<t})$ by masking $S_t$ to yield scores $\tilde{S}$ that satisfy (1-5) as follows:
\begin{enumerate}
    \item $\tilde{S}_{\cdot, 0}=-\infty$
    \item $A_{i,j}=1$ implies $\tilde{S}_{\cdot, j}=-\infty$
    \item $A_{i,j}=1$ implies $\tilde{S}_{i,j}=-\infty$
    \item $\tilde{S}_{i,i}=-\infty$ for all $i$
    \item $R_{i,j}=1$ implies $\tilde{S}_{j,i}=-\infty$, where $R\in \{0,1\}^{N\times N}$ is the reachability matrix (transitive closure) of $A$. That is, $R_{i,j}=1$ when there is a directed path from $i$ to $j$. \footnote{The reachability matrix $R$ can be computed with batched matrix multiplication as $\sum_{k=1}^t A^k$ where $t$ is the maximum path length; other methods could potentially improve speed.}
\end{enumerate}

The selected edge is then $\arg\max_{(i,j)} \tilde{S}_{i,j}$.

A full tree is decoded by calling $f_{\text{valid}}$ for $T$ steps, using the current step scores $S_t$ and an adjacency matrix $A_{<t}=\bigcup_{t'=1}^{t-1} \{(i,j)_{t'}\}$.

\paragraph{Multiple Edges Per Step} To decode multiple edges per step, i.e. $|E_t|\geq1$, we propose to repeatedly call $f_{\text{valid}}$, adding the returned edge to the adjacency matrix after each call, and stopping once the returned edge's score is below a pre-defined threshold $\tau$. %We save more sophisticated strategies as future work.

% \subsection{Additional Results}
% \begin{figure*}
% \includegraphics[height=0.60\textheight]{images/unif.png}
% \caption{Per-step edge distributions $\pi_{\theta}((i,j)|\hat{E}_{<t},x)$ on a 20-word example from a recurrent weight model trained with a \textbf{uniform oracle}. Each box shows a distribution, along with the edge predicted at that step using a greedy valid decoder. The model aims to place equal mass on each free edge $(i,j)\in \hat{E}^t_{\text{free}}$, and can adjust a distribution based on previous predictions. For instance, at $t=1$, edge $(19\rightarrow 17)$ is ranked lower than edge $(5\rightarrow 20)$, but not at $t=4$.}
% \end{figure*}
% \begin{figure*}
% \includegraphics[height=0.60\textheight]{images/coach.png}
% \caption{Per-step edge distributions $\pi_{\theta}((i,j)|\hat{E}_{<t},x)$ from a recurrent weight model trained with a \textbf{coaching oracle} using annealing. The model has learned to prefer certain free edges over others at each step. The low entropy distributions at early timesteps followed by higher entropy distributions later on (e.g. $t\in\{9,10,16\}$) may indicate easy-first behavior (taking entropy as an indicator of difficulty).}
% \end{figure*}
\end{document}